\documentclass{llncs}
\usepackage{makeidx}  
\usepackage{graphicx}
\usepackage{url}
\usepackage{pdfpages}
\usepackage{amsmath}
\usepackage{algorithm}
\usepackage[noend]{algpseudocode}
\usepackage{amsfonts}
\usepackage{bm}
\usepackage{multirow}
\usepackage{booktabs}
\usepackage{color}
\usepackage[font=footnotesize,labelfont=bf]{caption}
\usepackage{fancyhdr}
\usepackage{lastpage}

\makeatletter
\def\BState{\State\hskip-\ALG@thistlm}
\makeatother

\begin{document}
%
%
%
\title{Seq2RDF: An end-to-end application for deriving Triples from Natural Language Text}

\titlerunning{Seq2RDF }  
%
\author{Yue Liu, Tongtao Zhang, Zhicheng Liang,  Heng Ji, Deborah L. McGuinness}
\authorrunning{Liu et al.} 

\institute{Department of Computer Science, Rensselaer Polytechnic Institute}

\maketitle      
\vspace{-5mm}
\begin{abstract}
We present an end-to-end approach that takes unstructured textual input and generates structured output compliant with a given vocabulary. We treat the triples within a given knowledge graph as an independent graph language and propose an encoder-decoder framework with an attention mechanism that leverages knowledge graph embeddings. Our model learns the mapping from natural language text to triple representation in the form of {\tt subject-predicate-object} using the selected knowledge graph vocabulary. Experiments on three different data sets show that we achieve competitive F1-Measures over the baselines using our simple yet effective approach. A demo video is included.
\end{abstract}

\section{Introduction}
\label{introduction}
Converting free text into usable structured knowledge for downstream applications usually requires expert human curators, or relies on the ability of machines to accurately parse natural language based on the meanings in the knowledge graph (KG) vocabulary. Despite many advances in text extraction and semantic technologies, there is yet to be a simple 
system that generates RDF triples from free text given a chosen KG vocabulary in \textit{just one step}, which we consider an end-to-end system. 
We aim to automate the process of translating a natural language sentence into a structured triple representation
defined in the form of {\tt subject-predicate-object, s-p-o} for short, and build an end-to-end model based on an encoder-decoder architecture that learns the semantic parsing process from text to triple without tedious feature engineering
and intermediate steps.
We evaluate our approach on three different datasets and achieve competitive F1-measures 
outperforming our proposed baselines, respectively. The system, data set and demo are publicly available\footnote{\url{https://github.com/YueLiu/NeuralTripleTranslation}}\footnote{\url{https://youtu.be/ssiQEDF-HHE}}.
\vspace*{-2mm}
\section{Our Approach}

Inspired by the sequence-to-sequence model\cite{sutskever2014sequence} in recent Neural Machine Translation, we attempt to use this model to bridge the gap between natural language and triple representation. We consider a natural language sentence $X = [x_1, \ldots, x_{|X|}]$ as a source sequence, and we aim to map $X$ to an RDF triple $Y = [y_1, y_2, y_3]$ with regard to {\tt s-p-o} as a target sequence that is aligned with a given KG vocabulary set or schema. Given DBpedia for example, we take a large amount of existing triples from DBpedia as ground truth facts for training. 
Our model learns how to form a compliant triple with appropriate terms in the existing vocabulary. Furthermore, the architecture of the decoder enables the model to capture the differences, dependencies and constraints when selecting {\tt s-p-o} respectively, which makes the model a natural fit for this learning task.\\
\vspace{-5mm}
\begin{figure}[H]
\vspace{-5mm}
\centering
\includegraphics[width=11.2cm]{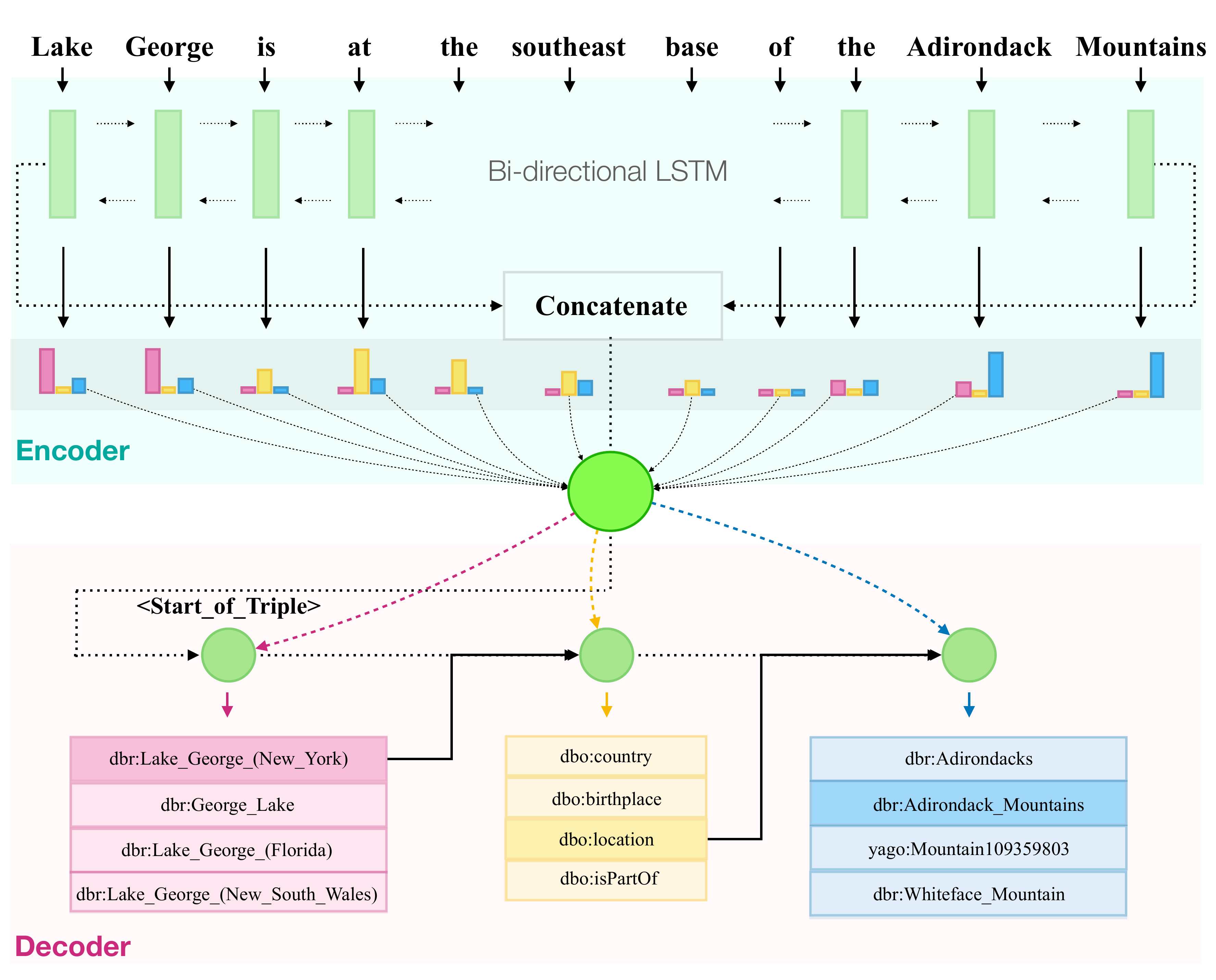}
\caption{Model Overview. Three colors (red, yellow, blue) represent the active attention during {\tt s-p-o} decoding respectively. We currently only generate a single triple per sentence, leaving the generation of multiple triples per sentence for future work.}
\label{fig:Seq2Seq}
\end{figure}
\vspace{-5mm}
\noindent
As shown in Figure~\ref{fig:Seq2Seq},
the model consists of an encoder taking in a natural language sentence as sequence input and a decoder generating the target RDF triple. The model pursues the maximized conditional probability
\begin{equation}
\label{eqn:formulation}
p(Y|X) = \prod_{t_d=1}^3p(y|y_{<t_d}, X),
\end{equation}
Both encoder and decoder are recurrent neural networks\footnote{We use {\tt tf.contrib.seq2seq.sequence\_loss} which is a weighted cross-entropy loss for a sequence of logits. We concatenate the last hidden output of forward and backward LSTM networks, the concatenated vector comes with fixed dimensions} with Long Short Term Memory (LSTM) 
cells.
We apply the attention mechanism that forces the model to learn to focus on specific parts of the input sequence when decoding, instead of relying only on the last hidden state of the encoder. Furthermore, in order to 
capture the semantics of the entities and relations within our training data, 
we apply domain specific resources\cite{liu2015exploiting} to obtain the word embeddings and the TransE model\cite{bordes2013translating} to obtain KG embeddings for entities and relations in the KG. We use these pre-trained Word embeddings and KG embeddings for entities and relations to initialize the encoder and decoder embedding matrix, respectively, and results show that 
this approach improves the overall performance.
\vspace*{-2.5mm}
\section{Experiments}

\textbf{Data Sets} We 
ran experiments on two public datasets {\tt NYT\footnote{New York Times articles: \url{https://github.com/shanzhenren/CoType}}\cite{ren2017cotype}}, {\tt ADE\footnote{Adverse drug events: \url{https://sites.google.com/site/adecorpus}}} with selected vocabularies and a {\tt Wiki-DBpedia} dataset that is produced by distant supervision\footnote{\url{http://deepdive.stanford.edu/distant_supervision}}. For data obtained by distant supervision, the test set is manually labeled to ensure its quality. 
Each data set is an annotated corpus with corresponding triples in the form of either {\tt s-p-o} or {\tt entity mentions} and {\tt relation types} at the sentence level. Details are available on our 
GitHub page.
\vspace*{-3mm}
\begin{table}[H] 
\vspace*{-3mm}
\centering
\begin{tabular}{llp{6cm}}
\hline
 \textbf{Text} &  &Berlin is the capital city of Germany.\\
 \textbf{Triple} &  & {\tt dbr:Germany dbo:capital dbr:Berlin}\\
 \hline
\end{tabular}
\caption{Example annotated pair with distant supervision on Wiki-DBpedia}
\label{dis-sup}
\end{table}
\vspace*{-9mm}
\noindent
\textbf{Evaluation Metrics}
We consider pipeline-based approaches that combine Entity Linking (EL) and Relation Classification (RC) as 
state of the art. 
We propose several baselines with combined outputs from state-of-the-art EL\footnote{Stanford, Domain specific NER} and RC for evaluation. We use F1-measure to evaluate triple generation (an output is considered correct only if {\tt s-p-o} are all correct) in comparison with the baselines.\\

\noindent
\textbf{Baselines} We implement multiple baselines including a classical supervised learning using simple Lexical features, a state-of-the-art recurrent neural network (RNN) approach with LSTM~\cite{miwa2016end} and one with 
a Gate Recurrent Unit (GRU) variant.
Then we evaluate the performance on triple generation with results combining EL and RC. The hyper-parameters in our model are tuned with 10-fold cross-validation on the training set according to the best F1-scores. We applied the same settings to the baselines. The details regarding the parameters and settings are available on our 
GitHub
page for replication purposes.
\vspace*{-2mm}
\section{Result Analysis}
We achieve 
the best F1 Measure of 84.3 on the triple generation from Table~\ref{f1-whole}. Note that the baseline approaches that we implemented are pipeline-based, and thus they are very likely to propagate errors to downstream components. However, our model merges the two different tasks of EL and RC into one during the decoding, which composes a major advantage over pipeline-based approaches that usually apply separate models on EL and RC. The most common errors are caused by {\tt Out of vocabulary} and {\tt Noise from overlapping relations} in text. As we do not cover all rare entity names or consider multiple triple situations, these errors are valid in some sense.
\vspace*{-2mm}
\begin{table}[H]
\vspace*{-2mm}
\centering
\begin{footnotesize}
\begin{tabular}{lccc}
\toprule
Tasks & \multicolumn{1}{c}{NYT}  & \multicolumn{1}{c}{ADE} & \multicolumn{1}{c}{Wiki-DBpedia}  \\
\cmidrule(l){2-2}\cmidrule(l){3-3}\cmidrule(l){4-4}
 Metric & F1-Measure  & F1-Measure & F1-Measure \\
 \midrule
 EL+Lexical & 36.8 & 61.4 & 37.8  \\
 EL+LSTM & 58.7  & 70.3 & 65.5\\
 EL+GRU & 59.8  & 73.2 & 67.0\\
 Seq2Seq& 64.2  & 73.4 & 73.5\\
 S+A+W+G & \textbf{71.4}  & \textbf{79.5} & \textbf{84.3}\\
\bottomrule
\end{tabular}
\end{footnotesize}
\caption{Cross-dataset comparison on triple generations. Seq2Seq denotes the implementation of Seq2Seq without any attention mechanism and pre-trained embeddings; A denotes attention mechanism; W and G denote pre-trained word embeddings for the encoders and KG embeddings for the decoders, respectively.}
\label{f1-whole}
\end{table}
\vspace{-14.3mm}
\section{Conclusions and Future Work}
We present an end-end system for translating a natural language sentence to its triple representation. Our system performs competitively on three different datasets and our assumption on enhancing the model with pre-trained KG embeddings improves performance across the board. It is easy to replicate our work and use our system following the demonstration. In the future, we plan to redesign the decoder and enable the generation of multiple triples per sentence.\\
\textbf{Acknowledgement} This work was partially supported by the NIEHS Award 0255-0236-4609 / 1U2CES026555-01.
\vspace{-2mm}
\bibliography{refs}
\bibliographystyle{splncs03}

\end{document}